\documentclass[review]{elsarticle}
\graphicspath{ {./figures/} }
\usepackage{hyperref}
\usepackage{float}
\usepackage{verbatim} 
\usepackage{apalike}
\usepackage{amsmath,amssymb,amsfonts}
\usepackage{algorithmic}
\usepackage{graphicx}
\usepackage[table,xcdraw]{xcolor}
\usepackage{multirow}
\usepackage{array}
\usepackage{booktabs}
\usepackage{caption}
\usepackage{textcomp}
\usepackage{nicematrix}

\usepackage{algorithmic}
\usepackage[ruled,vlined,linesnumbered,boxruled]{algorithm2e}
\usepackage{setspace}
\usepackage{xcolor}
\usepackage{nicematrix}
\usepackage{mfirstuc}
\MFUnocap{of}
\MFUnocap{or}
\MFUnocap{etc}
\MFUnocap{are}
\MFUnocap{if}
\MFUnocap{or}
\MFUnocap{on}
\MFUnocap{in}

\restylefloat{figure}
\restylefloat{table}

\journal{Expert Systems with Applications}

\biboptions{authoryear}

\begin{document}
\begin{frontmatter}

\begin{titlepage}
\begin{center}
\vspace*{1cm}


\textbf{ \large {A Framework for Generalizing Critical Heat Flux Detection Models Using Unsupervised Image-to-Image Translation}}

\vspace{1.5cm}

Firas Al-Hindawi$^{a}$ (falhinda@asu.edu), Tejaswi Soori$^b$ ({sooritejaswi@gmail.com}), Han Hu$^c$ (hanhu@uark.edu), {Md Mahfuzur Rahman Siddiquee$^a$ (mrahmans@asu.edu)}, Hyunsoo Yoon$^d$ (hs.yoon@yonsei.ac.kr), Teresa Wu$^a$ (teresa.wu@asu.edu), Ying Sun$^e$ ({sunyg@ucmail.uc.edu}) \\

\hspace{10pt}

\begin{flushleft}
\small  
$^a$ Arizona State University, 699 Mill Avenue, Tempe, AZ 85281, US\\
$^b$ Drexel University, 3141 Chestnut Street, Philadelphia, PA 19104, US\\
$^c$ University of Arkansas, 1 University of Arkansa, Fayetteville, AR 72701, US\\
{$^d$   Yonsei University, 50 Yonsei-ro, Seodaemun-gu, Seoul, 03722, South Korea}\\
{$^e$  University of Cincinnati, 2600 Clifton Ave, Cincinnati, OH 45221, US}\\

\vspace{1cm}

\textbf{Corresponding Author:} \\
Firas Al-Hindawi \\
Arizona State University, 699 Mill Avenue, Tempe, AZ 85281, US \\
Tel: +1 (602) 837-9820 \\
Email: falhinda@asu.edu

\end{flushleft}        
\end{center}
\end{titlepage}

\title{{ A Framework for Generalizing Critical Heat Flux Detection Models Using Unsupervised Image-to-Image Translation}}

\author[label1]{Firas Al-Hindawi \corref{cor1}}
\ead{falhinda@asu.edu}

\author[label2]{Tejaswi Soori}
    \ead{sooritejaswi@gmail.com} 
    
\author[label3]{Han Hu}
    \ead{hanhu@uark.edu}

\author[label1]{{Md Mahfuzur Rahman Siddiquee}}
    \ead{mrahmans@asu.edu}

    
\author[label4]{Hyunsoo Yoon}
    \ead{hs.yoon@yonsei.ac.kr}
    
\author[label1]{Teresa Wu}
    \ead{teresa.wu@asu.edu}
    
\author[label5]{Ying Sun}
    \ead{sunyg@ucmail.uc.edu}

\cortext[cor1]{Corresponding author.}
\address[label1]{Arizona State University, 699 Mill Avenue, Tempe, AZ 85281, US}
\address[label2]{Drexel University, 3141 Chestnut Street, Philadelphia, PA 19104, US}
\address[label3]{University of Arkansas, 1 University of Arkansas, Fayetteville, AR 72701, US}
\address[label4]{{Yonsei University, 50 Yonsei-ro, Seodaemun-gu, Seoul, 03722, South Korea}}
\address[label5]{{University of Cincinnati, 2600 Clifton Ave, Cincinnati, OH 45221, US}}

\begin{abstract}

{ The detection of critical heat flux (CHF) is crucial in heat boiling applications as failure to do so can cause rapid temperature ramp leading to device failures. Many machine learning models exist to detect CHF, but their performance reduces significantly when tested on data from different domains. To deal with datasets from new domains a model needs to be trained from scratch. Moreover, the dataset needs to be annotated by a domain expert. To address this issue, we propose a new framework to support the generalizability and adaptability of trained CHF detection models in an unsupervised manner. This approach uses an unsupervised Image-to-Image (UI2I) translation model to transform images in the target dataset to look like they were obtained from the same domain the model previously trained on. Unlike other frameworks dealing with domain shift, our framework does not require retraining or fine-tuning of the trained classification model nor does it require synthesized datasets in the training process of either the classification model or the UI2I model. The framework was tested on three boiling datasets from different domains, and we show that the CHF detection model trained on one dataset was able to generalize to the other two previously unseen datasets with high accuracy. Overall, the framework enables CHF detection models to adapt to data generated from different domains without requiring additional annotation effort or retraining of the model.}
\end{abstract}


\end{frontmatter}

\section{Introduction}
\label{introduction}

Boiling is a heat transfer mechanism that dissipates a large amount of heat with minimal temperature increase by taking the advantage of the latent heat of the working fluid. As such, boiling has been widely implemented. Nevertheless, the heat flux of boiling is bounded by a practical limit known as the critical heat flux (CHF), beyond which, a continuous vapor layer will blanket the heater surface, leading to a significant reduction in heat transfer coefficient (HTC) and deteriorating the heat dissipation; this process is also known as the critical heat flux (a.k.a. boiling crisis). Upon the occurrence of CHF, a rapid temperature ramp takes place on the heater surface which will lead to detrimental device failures.  
To capture CHF, a variety of theoretical models have been developed based on different transport mechanisms during boiling, including hydrodynamic instabilities \citep{zuber1959a}, force balance \citep{kandlikar2001a} , etc. However, due to the complexity and stochasticity of the boiling process, existing theoretical CHF models are overly-simplified to accurately predict CHF before it happens. Often, a safety factor (e.g. the departure from the nucleate boiling ratio (DNBR) in nuclear reactors) is applied to avoid the boiling crisis, leading to reduced system performance \citep{lee2021a}. 

Boiling images have attracted great attention lately as the images contain detailed information on the bubble dynamics. It is of both fundamental and practical interest to detect the CHF and identify the dominant transport mechanism that triggers CHF using boiling images. In the side view boiling images, bubbles at different locations are at different stages of the ebullition cycle. As such, under steady-state conditions, the dynamics of the bubbles are embedded in static boiling images. A static image of pool boiling may cover the entire ebullition cycle, including bubble nucleation, growth, and departure. Traditional analysis of the boiling images relies on the extraction of physical parameters based on domain knowledge, such as bubble size, bubble departure frequency \citep{li2019a}, nucleation site density \citep{park2016a}, void fraction \citep{ridwan2019a}, vapor film \citep{allred2018a}, etc. But the stochasticity of the bubble dynamics adds to the fluctuations and uncertainties of the extracted physical parameters. Furthermore, the traditional boiling image analysis is limited to known physical parameters, whereas it is not clear whether the known parameters are sufficient to capture the boiling crisis.

Machine learning algorithms have been widely used in various engineering applications \citep{altarazi2019machine,alhindawi2018predicting,rokoni2022a,rassoulinejad-mousavi2021a,zhao2022subdomain,ji2022machine, wang2022simultaneous}. Specifically for CHF, researchers started developing models using a variety of supervised learning algorithms, including support vector machine \citep{hobold2018a}, multilayer perceptron (MLP) neural networks  \citep{hobold2018a}, and convolutional neural networks (CNN) \citep{rassoulinejad-mousavi2021a}, and using different modalities such as acoustic emissions \citep{sinha2021a}, optical images \citep{rokoni2022a}, and thermographs \citep{ravichandran2021a} to predict boiling heat flux or/and the boiling regime. While these studies have shown high prediction accuracy and success in heat flux detection, most of them were only trained and tested on a single-source dataset, e.g., the authors’ own experimental data. For applications where there is a domain shift problem (the target data are drawn from a different distribution than the source training data), the performance of the model declines dramatically and in extreme cases, it may become worse than random guessing\citep{wilson2020a}.

{Recent efforts were dedicated to alleviating this problem. Transfer learning (TL) was used to adapt a trained CNN model for boiling regime classification to a new target domain with a small volume of labeled data from the target domain \citep{rassoulinejad-mousavi2021a}. By taking features from the trained CNN model and fine-tuning the networks, the TL model requires much less labeled data from the target domain than CNN to yield the same level of prediction accuracy. Nevertheless, this TL approach still relies on labeled data from the target domain and is not applicable to an unlabeled target domain.

The utilization of GANs and unsupervised domain adaptation methods to solve the domain shift in heat transfer processes has been limited with several novel applications in studying fluid dynamics problems only such as unsteady flows\citep{lee2019a,deng2019a}, turbulence closure models\citep{bode2021a}, and flow field visualization\citep{lee2019a}. Please note these approaches are not  designed to support unsupervised cross-domain classification (e.g., data collected from different sources). There has yet to emerge a framework to support the unsupervised generalizability and adaptability of CHF detection models for the boiling crisis problem which is the goal of this research.}

{ In this research, we propose a new framework to generalize a pre-trained CHF detection model and enable it to adapt to data generated from a different domain. The framework consists of two parts. The first part utilizes a pre-trained typical classification model that is trained and tested on the source dataset. The second part utilizes an Unsupervised Image-to-Image (UI2I) translation model to transform images in the target dataset to look as if they were obtained from the same domain of the source dataset. Instead of spending resources on manually labeling each new dataset and building separate classification models for each one, we can adapt an existing classification model to new datasets by incorporating them into the familiar domain, without requiring human supervision.  

A total of three datasets from different domains were used. Two of which were publicly available datasets (DS-1 and DS-2) and the third was obtained from in-house experiments (DS-3). We used the public dataset with the lowest resolution (DS-1) as our source dataset and alternated DS-2 and DS-3 with higher resolution images as the target dataset. This demonstrates the ability of the framework to work even from low resolution to high resolution and when using publicly available datasets as the base. Determining whether two domains are transferable in unsupervised image-to-image translation applications can be a challenging task. Typically, this task involves utilizing techniques like visual inspection, statistical analysis, or domain expertise. Often a combination of such methods is required. In our case, we relied on visual inspection and domain knowledge to make this decision. The datasets employed in this work depict boiling images conducted under different circumstances exhibiting the same physical phenomena that accompany the boiling heat transfer mechanism. This resemblance can be spotted by domain expert' naked eyes.   


It is worth mentioning that there  has been a number of approaches that leverage GANs and UI2I models to address the bias between different domains. For example, \cite{deng2018image} translated the source DS to the target domain and then trained a new model on the features of the translated images. In \cite{xiang2020unsupervised}, the authors synthesized a dataset and generated different contextual conditions on the synthetic data set. They created labels for the synthesized dataset  and fine-tuned the model using the synthesized dataset. In general, our method shares an inherited sub-problem with these approaches developed to address the domain shift challenges. However, our solutions fundamentally differ. Specifically, of particular interest in our study is the generalizability of the classification-based framework which can adopt unlabeled data collected from other sources without the need of extra efforts such as fine-tuning as in \cite{xiang2020unsupervised} or even training a new classifier as in \cite{deng2018image}. We contend this retrain-free or fine-tuning-free approach is scalable to applications with multiple domains.}

{
To summarize, the contribution of this paper includes: 
\begin{itemize}
\item Introducing a framework to support the generalizability and adaptability of CHF detection models to unlabeled datasets coming from domains never seen previously by the model in a fully unsupervised manner using a UI2I translation model.
\item The framework was applied to generalize a CNN classification model, but it is model agnostic and could be applied to any type of classification models. 
\item The framework used FP-GAN for UI2I translation, but it is agnostic to the UI2I model used. FP-GAN could be replaced by any other model. This means that the more the UI2I models advance, the better the performance of the framework.
\item Unlike other frameworks tackling the domain shift problem, our framework does not require re-training or fine-tuning of the trained classification model, nor does it require synthesized datasets in the training process of either the classification model or the UI2I model.
\end{itemize}
}


{The paper is organized as follows, we first discuss the related work in the literature. Next, we introduce the proposed framework in the Methods section. Afterward, in the experiments section, we discuss data preparation, the source classification model training, and the UI2I translation model training. We follow that with the results and discussion section where we discuss our findings and analysis of the results before we finally conclude the paper.}

{
\section{Related Work}

(GANs) were initially introduced by \cite{goodfellow2014a}. In their work, the authors proposed a model architecture that consists of two networks (a generator and a discriminator) that are trained together to generate synthetic data that resembles a given dataset. 


The first work to utilize GANs to solve the I2I translation problem was \cite{isola2017image}. In their pix2pix model, they used conditional adversarial networks to learn the mapping from an input image to an output image, where the networks learn a loss function to train this mapping. This method uses a "U-Net" based architecture for the generator and a "PatchGAN" classifier for the discriminator. 


The main limitation of the pix2pix model was that it was supervised. The training process required  paired images in the training set for the model to learn the mapping $G: X \rightarrow Y$, where generated images from $G(X)$ are indistinguishable from real images coming from domain $Y$. Simply using the adversarial loss for this problem makes it heavily under-constrained. Thus, to solve this problem, \cite{zhu2017unpaired} introduced one of the most famous unsupervised I2I models known as CycleGAN. The authors coupled the adversarial loss with an inverse mapping $F: Y \rightarrow X$ and introduced a cycle consistency loss. The objective of this loss is to enforce $F(G(X)) \approx X$ and $G(F(Y)) \approx Y$. The idea was inspired by the language translation process, where a reverse translation of a sentence that was translated from language A to language B should give the same original sentence in language A. A similar approach was performed by \cite{yi2017dualgan} and \cite{kim2017learning} concurrently with cycleGAN. \cite{li2018unsupervised} proposed the SCANs framework to address the shortcomings of UI2I models in handling problems where there is a marginal difference between the domains or when the images are of high resolution. Their framework works by decomposing a single translation into multi-stage transformations where the information from the previous stage is used in the next stage using an adaptive fusion block. \cite{kim2019u} incorporated a new attention module to guide the model in distinguishing between source and target domains by focusing on the most important features.

\cite{Choi_2018_CVPR}  introduced their StarGAN framework that simultaneously trains multiple datasets with different domains using a single generator and discriminator pair. However, StarGAN tends to change the images unnecessarily during image-to-image translation even when no translation is required \citep{siddiquee2019a}. To address this issue, \cite{siddiquee2019a} proposed the Fixed-Point GAN (FP-GAN) framework. This framework focused on identifying a minimal subset of pixels for domain translation and introduced fixed-point translation by supervising same-domain translation through a conditional identity loss, and regularizing cross-domain translation through revised adversarial, domain classification, and cycle consistency loss.

GAN-based I2I translation frameworks have been widely applied in a variety of fields. In the medical imaging field for example, \cite{amirkolaee2022development} proposed a novel GAN for medical I2I translation to enhance the quality of medical imaging for diagnostic and therapeutic purposes. \cite{arruda2022cross} proposed a novel multidomain signal-to-signal translation method using a StarGAN-based model to generate artificial steady-state visual evoked potential (SSVEP) signals from resting electroencephalograms to improve the performance of SSVEP-based brain-computer interfaces. In the field of autonomous driving, \cite{Tremblay_2018_CVPR_Workshops} employed I2I translation models to support the training of autonomous driving algorithms. In the field of text extraction, \cite{kundu2020text} utilized I2I transltion to address the problem of text-line extraction (TLE) from unconstrained handwritten document images. \cite{deng2018image,xiang2020unsupervised} utilized UI2I to solve person re-identification problem. In the area of fluid dynamics, \cite{deng2019a} used GANs to reconstruct high-fidelity data from low-fidelity Particle Image Velocimetry (PIV) data for flow around single and multiple cylinders \citep{deng2019a}. \cite{lee2019a} predicted unsteady flow around a cylinder by using the GAN to reproduce flow fields that match Large Eddy Simulations (LES). 

Unfortunately, albeit all this, the utilization of GANs and unsupervised domain adaptation methods to study heat transfer processes has been rather limited and the specific use of UI2I translation models is even more scarce. One major challenge preventing this is the lack of high-quality labeled data for training and evaluation. Collecting and labeling data for boiling crisis and CHF detection is a difficult and expensive process, as these phenomena occur under extreme conditions and require specialized equipment and expertise for measurement. Furthermore, the lack of a clear understanding of the underlying mechanisms of boiling crisis and CHF detection limits the development of accurate models. These phenomena are still not fully understood, and the complexity of the phenomena makes it challenging to develop models that embed the underlying physics in the model.




}

\begin{figure}[!t]
\captionsetup{justification=centering}
\centerline{\includegraphics[width=\columnwidth]{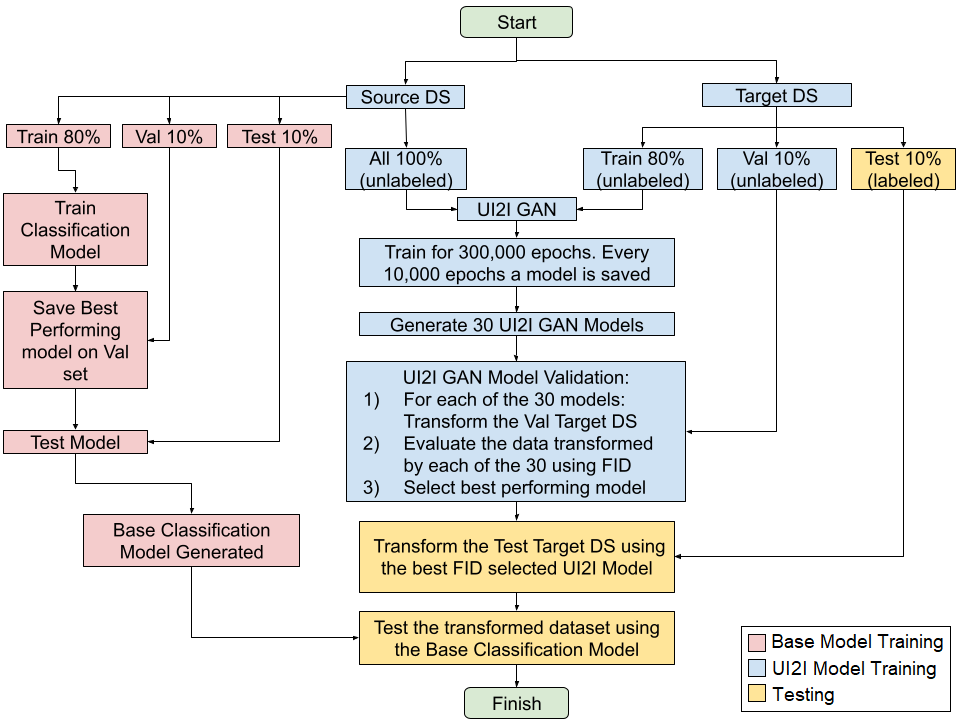}}
\caption{\capitalisewords{Framework Summary Flow Chart: the red part of the flow chart represents the source classification model training process. The blue part represents the UI2I translation model training process. The yellow part represents the testing process.}}
\label{fig1}
\end{figure}

{
\section{Methods}
The proposed framework is summarized in Fig\ref{fig1}. The framework consists of two components, the first component utilizes a typical classification model that is trained and tested on the source dataset. This process is depicted in red in the figure. The specific details of the source classification model training are discussed further in section \ref{Source Classification Model Training}. The second component utilizes a UI2I translation model to transform images in the target dataset to look as if they were obtained from the same domain of the source dataset. The model was trained 300k iterations and a checkpoint model was saved every 10k iterations. After the UI2I training was finished, Validation was conducted on all checkpoint models saved and the best-performing model was move forward for final testing. The UI2I training process is depicted in blue in the figure and the final testing is depicted in yellow. The specific details of the UI2I translation model training and testing are discussed further in section \ref{UI2I Translation Model Training}. The method is further described in the pseudo-code in Algorithm \ref{alg:Pseudo code}.
}

This way, instead of generalizing the classification model by either training it on hundreds of labeled datasets (which is infeasible) or by training a new model for every new experiment, we can use this framework to make the data adapt to a pre-existing model. Although we are using this framework on boiling image datasets, the concept could be easily translated to any case scenario where there exists a pre-trained model on one domain that needs to be generalized to datasets from other domains.

\begin{algorithm}[!t]

\scriptsize 
\setstretch{0.6}
\caption{Framework pseudo code}

\label{alg:Pseudo code}
  \DontPrintSemicolon
    \SetKwFunction{one}{train\_source\_classifier}
    \SetKwFunction{two}{train\_UI2I\_model}
    \SetKwFunction{three}{translate\_target}
    \SetKwFunction{four}{validation\_and\_testing}
  
  \SetKwProg{Fn}{Function}{:}{}

 \textbf{ Step 1:}
  \Fn{\one{$source\_DS$}}{
        train model\;
        save best model\;
        source-DS classifier = best saved model\;
        \KwRet{source-DS classification model}\;
  }
  
  \textbf{Step 2:}
  \Fn{\two{$source\_DS$,$target\_DS$}}{
        \textbf{Initialize} models\_list = [ ]\;
        use source\_DS and target\_DS to train UI2I translation model\;
        run training for $300 k$ iterations\;
        save a checkpoint model every $10 k$ iterations\;
        append saved checkpoint model to models\_list\;
        \KwRet{models\_list [$10k$-model, $20k$-model, ... , $300k$-model]}\;
  }
  \textbf{Step 3:}
    \Fn{\three{$target\_DS$,$models\_list$}}{
        \textbf{Initialize} translated\_sets\_list = [ ]\;
        \textbf{For} each model in models\_list:\;
        \Indp translate target\_DS to source domain using model\;
        append translated images set to translated\_sets\_list\;
        \Indm \KwRet{translated\_sets\_list [$source\_DS^*-10k$, $source\_DS^*-20k$, ... , $source\_DS^*-300k$]}\;
  }
  \textbf{Step 4:}
    \Fn{\four{$translated\_sets\_list$,$test\_set$}}{
        \textbf{Initialize} best\_FID, best\_translated\_set, best\_UI2I\_model\;
        \textbf{For} each translated\_set in translated\_sets\_list:\;
            \Indp \textbf{IF} FID(translated\_set) $<$ best\_FID:\;
                \Indp \Indp  best\_FID = FID(translated\_set)\;
                             best\_translated\_set= translated\_set\;
        \Indm \Indm \Indm  best\_UI2I\_model = UI2I of best\_translated\_set\;
        translate and evaluate test\_set using best\_UI2I\_model\;
  }

\end{algorithm}

\section{Experiments}
\subsection{Data Preparation}
Three different pool boiling experimental image datasets (DS-1, DS-2, and DS-3) were prepared in this study, where DS-1 and DS-2 were generated using publicly available YouTube videos \citep{you-a,minseok2014a} while DS-3 was conducted in-house. Specifically, the video from which DS-1 was prepared shows a pool boiling experiment performed using a square heater made of high-temperature, thermally-conductive microporous coated copper where the surface was fabricated by sintering copper powder. The square heater had a surface area of $\approx$ $100$ $mm^2$ and the working fluid used was water. All experiments were performed at a steady-state under an ambient pressure of 1 atm. A T-type thermocouple was used for temperature measurements. The resolution of the video frames was $512\times480$ pixels. The YouTube video from which DS-2 was prepared shows a pool boiling experiment performed using a circular heater made of microporous coated copper where the surface was fabricated by sintering copper powder. The circular heater had a diameter of $\approx$ $16$ $mm$ and the working fluid used was DI water. All experiments were performed at a steady state under an ambient pressure of 0.5 atm. A T-type thermocouple was used for temperature measurements. The resolution of the video frames was $1280\times720$ pixels. DS-3 was obtained from our in-house experiments of water boiling on polished copper surfaces with an area of $100$ $mm^2$. The high-speed videos were captured using Phantom VEO 710L at a frame rate of 3000 fps and a resolution of $1280 \times 800$ pixels.

Images for DS-1 and DS-2 were prepared by downloading the videos from YouTube and extracting individual frames using a MATLAB code via the VideoReader and imwrite functions. Recognizing duplicate frames extracted from the YouTube videos, quality control was conducted to remove the repeated images by calculating the relative difference using the Structural Similarity Index (SSIM) value \citep{gao2020a} between two consecutive images where images with a relative difference less than 0.03\% were removed. This pre-processing is important to ensure DL models were not biased by identical image frames. Benefiting from the large optical sensor size (25.6 mm × 16 mm) and the high-power backlight (Advanced Illumination BT200100-WHIIC), images of DS-3 have a balanced and homogeneous background. Also, the images are directly saved from the raw video files (.cine) that retain the highest image quality. As such, this pre-processing step was not necessary for DS-3. 

\begin{figure}[!t]
\captionsetup{justification=centering}
\centerline{\includegraphics[width = \columnwidth]{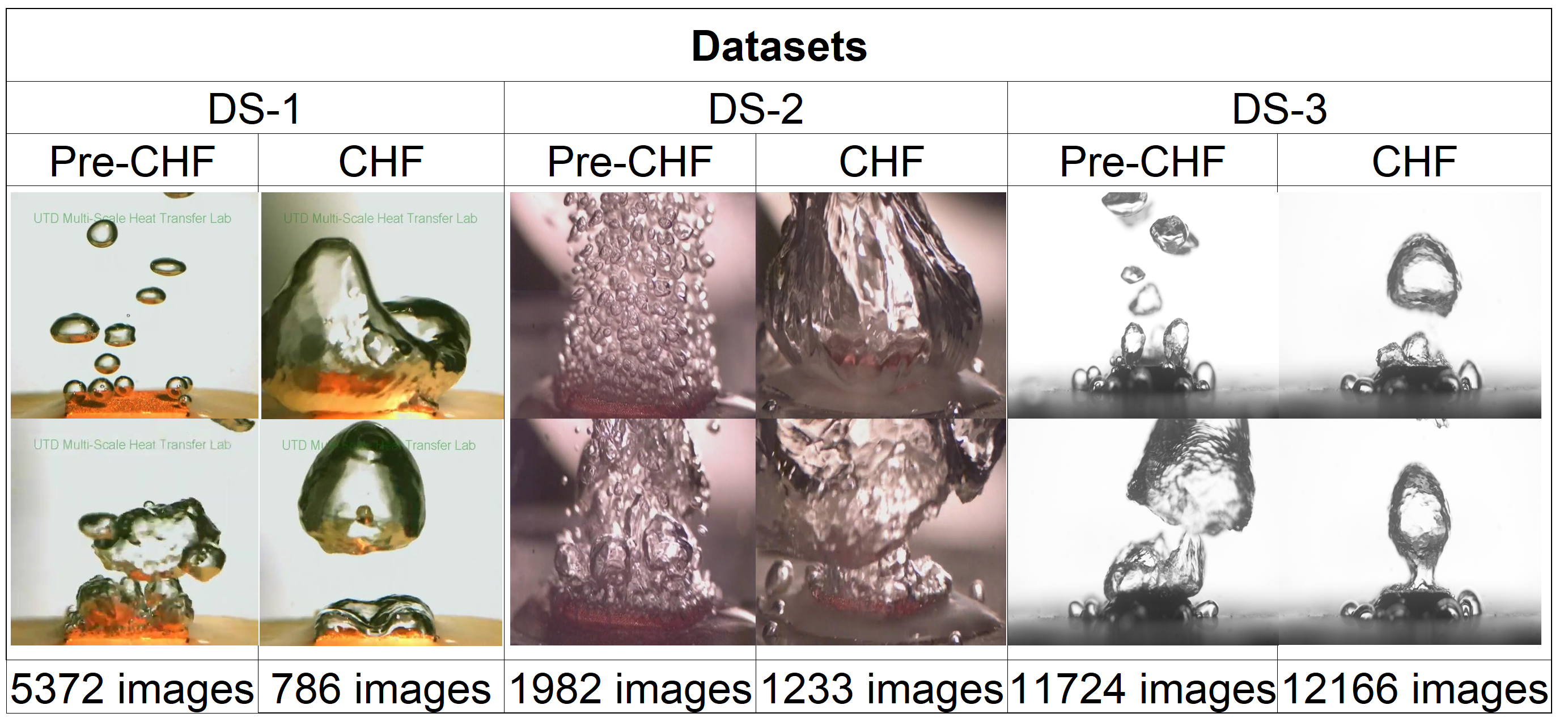}}
\caption{\capitalisewords{Representative Images of Bubble Dynamics From Source Videos.}}
\label{fig3}
\end{figure}

The images were categorized into two boiling regimes: (1) The critical heat flux regime (CHF), where a significant drop in the heat transfer coefficient is observed due to a continuous vapor layer blanketing the heater surface and (2) pre-CHF regime, where optimal heat transfer coefficient is obtained and  only discrete bubbles or frequent bubble coalescence is observed before departure. While images of DS-1 and DS-2 have been labeled by the authors of the datasets, this study is designed to be unsupervised learning thus labeling of DS-2 is only used to assess the model performance. Originally, DS-1 had a total of 6158 images (786 CHF versus 5372 pre-CHF), DS-2 had a total of 3215 (1233 CHF versus 1982 pre-CHF) and DS-3 had a total of 23890 (12166 CHF versus 11724 pre-CHF). As seen, all data sets were unbalanced. We used undersampling to balance DS-1. Datasets DS-2 and DS-3 were not balanced since the objective of this study is to introduce a framework that utilizes unsupervised learning, that is, the labeling information of DS-2 and DS-3 are assumed to be unavailable. Table\ref{table1} shows the number of images in each regime for each dataset before and after the down-sampling process and Fig\ref{fig3} shows a visual representation of the images for each dataset. The pixel intensity values in each image were normalized to fit in the range [0,1] to ensure uniformity over multiple datasets during deep learning training.

\begin{table}
\captionsetup{justification=centering}
\caption{Datasets Summary}
\label{table1}
\centering
\resizebox{\columnwidth}{!}{%
\begin{tabular}{>{\centering\arraybackslash}m{1cm}>{\centering\arraybackslash}m{2cm}>{\centering\arraybackslash}m{2cm}>{\centering\arraybackslash}m{2cm}>{\centering\arraybackslash}m{2cm}}
\toprule
 & \multicolumn{2}{c}{\textbf{Before under-sampling of DS-1}} & \multicolumn{2}{c}{\textbf{After under-sampling of DS-1}} \\
\cmidrule(lr){2-5}
\textbf{DS} & \textbf{CHF} & \textbf{Pre-CHF} & \textbf{CHF} & \textbf{Pre-CHF} \\
\midrule
DS-1 & 786 & 5372 & 786 & 786 \\
DS-2 & 1233 & 1982 & 1233 & 1982 \\ 
DS-3 & 12166 & 11724 & 12166 & 11724\\ 
\bottomrule             
\end{tabular}
}
\end{table}

{
\subsection{Source Classification Model Training} \label{Source Classification Model Training}
To demonstrate that the framework is agnostic to the classification model used, we tested three different model architectures. The first model is a customized architecture tied to the application and is summarized in Fig\ref{fig4}. The second and third model architectures are the famous ResNet50 \citep{He2015} and MobileNet\citep{howard2017mobilenets}. 

The dataset was divided into three parts: 1) a training set (80\%), 2) a validation set (10\%), and 3) a test set (10\%). Each model was trained for 100 epochs using an Adam optimizer and the model that scored the lowest loss on the validation set was saved to be used in our pipeline. Finally, the selected model was blind-tested on the test set for final evaluation.

\begin{figure}[!t]
\captionsetup{justification=centering}
\centerline{\includegraphics[width = \columnwidth]{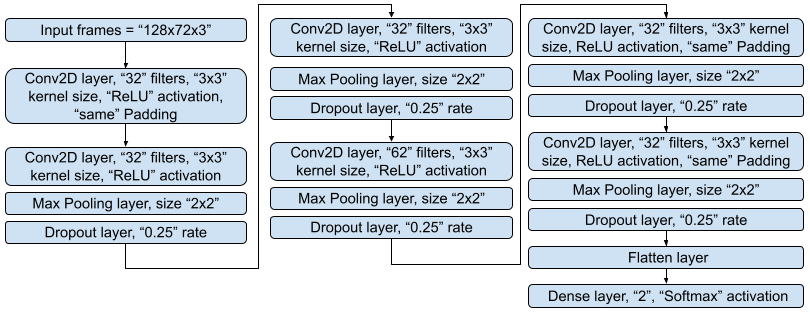}}
\caption{\capitalisewords{Architecture of The Customized Model.}}
\label{fig4}
\end{figure}
}

{
\subsection{UI2I Translation Model Training}\label{UI2I Translation Model Training}
The target dataset was divided into three parts: a) training set, b) validation set, and c) testing set. The training set from the target dataset was fed into the UI2I translation model along with 100\% of the source dataset. The UI2I model was trained for $300k$ Iterations, where for every $10k$ iterations a model was saved, making a total of $30$ UI2I translation models saved. For our implementation, we used the state-of-the-art FP-GAN \citep{siddiquee2019a} as the UI2I translation model of choice. FP-GAN was designed to preserve important features of the input image, such as edges, textures, and colors while maintaining semantic consistency. This is in contrast to other models which can sometimes lose important input features during the translation process. Moreover, it has been shown to produce high-quality results for a variety of image-to-image translation tasks, making it well-suited for our framework. That being said, We contend that the framework is agnostic to the UI2I model used and could be replaced by other UI2I translation models. For the training of FP-GAN, We used the same training settings as used by the authors and proved to be successful. 
}

To evaluate the UI2I translation model performance, we used the Fréchet Inception Distance (FID) metric. The FID reflects on the difference between two sets of images in terms of statistical vision features. Let $p(.)$ be the distribution of the InceptionV3 model internal representations (activations) of the images generated by the model and $p_w(.)$ the distribution of the same neural network activations from the "world" of real images used to train the model. The FID is basically the difference of the two Gaussians measured by the Fréchet distance also known as the Wasserstein-2 distance. The Fréchet distance $d(.,.)$ between the Gaussian with mean and covariance$(m,C$) obtained from $p(.)$ and the Gaussian $(m_w,C_w)$ obtained from $p_w(.)$ is called the FID.

$$
\begin{aligned}
\mathrm{FID} &=d^{2}\left((\boldsymbol{m}, \boldsymbol{C}),\left(\boldsymbol{m}_{\boldsymbol{w}}, \boldsymbol{C}_{\boldsymbol{w}}\right)\right) \\
&=\left|\boldsymbol{m}-\boldsymbol{m}_{\boldsymbol{w}}\right|_{2}^{2}+\operatorname{Tr}\left(\boldsymbol{C}+\boldsymbol{C}_{\boldsymbol{w}}-2\left(\boldsymbol{C} \boldsymbol{C}_{\boldsymbol{w}}\right)^{1 / 2}\right)
\end{aligned}
$$

The lower the score, the better the translation quality of the generated images by the UI2I translation model. The target validation data was fed into the 30 models to generate synthetic data images that look like source data images. Thus, a total of 30 sets of generated images were to be evaluated. The FID score was used to measure the distance between each of the 30 sets and the real source data images. The model that scored the lowest FID score on the validation set was the model to be selected to translate images from the target domain to the source domain. In theory, the framework process is at an end after selecting the best UI2I translation model to be deployed in production using the FID metric. However, since we do have the labels available, we utilized these labels to see if the framework was in fact doing what it was supposed to be doing in the final testing stage. In this part, we used the source classification model to find the actual best attainable UI2I translation model out of the saved 30 and then compare it with the best FID-selected UI2I translation model that was obtained without using the labels. After using the 30 models to generate the 30 newly translated datasets from the same validation set, the source classification model was used to classify the newly generated images and the results were evaluated using traditional supervised learning metrics to identify the best attainable UI2I translation model. Afterward, both the best attainable model and the best FID-selected model were used to translate the test data set and then classify both translated sets using the source classification model and finally evaluate it to further confirm the results.

\section{Results and Discussion}

\subsection{Experiment I: Source Classification Model and Blind Cross-Domain Testing}

In this section, the results obtained from the source classification models training are presented. Moreover, the results will be compared by blindly testing the source classification models on the target images directly (without translation). As seen in Table\ref{table2}, while the source classification models has satisfactory performance on the source DS on which the model was trained and validated, its performance on the target DS, which is previously unseen by the model is far from being acceptable as expected. This supports our efforts to explore the use of UI2I translation to improve the prediction of unseen data from different sources.

\begin{table}
\captionsetup{justification=centering}
\caption{\capitalisewords{Test Results of Source Classification Models on DS-1, DS-2 and DS-3} }
\label{table2}
\centering
\resizebox{\columnwidth}{!}{
\begin{tabular}{>{\centering\arraybackslash}m{2cm}>{\centering\arraybackslash}m{1cm}>{\centering\arraybackslash}m{2cm}>{\centering\arraybackslash}m{2cm}>{\centering\arraybackslash}m{2cm}>{\centering\arraybackslash}m{2cm}>{\centering\arraybackslash}m{2cm}}
\toprule
\textbf{Classifier} & \textbf{Test DS} & \textbf{Balanced Accuracy} & \textbf{F1 weighted} & \textbf{Precision weighted} & \textbf{Recall weighted} & \textbf{ROC AUC} \\
\midrule
  & DS-1 & 0.99 & 0.99 & 0.99 & 0.99 & 1.00 \\
 Customized Model & DS-2 & 0.50 & 0.47 & 0.38 & 0.62 & 0.37 \\ 
  & DS-3 & 0.50 & 0.32 & 0.24 & 0.49 & 0.89\\ 
  \midrule
  & DS-1 & 1.00 & 1.00 & 1.00 & 1.00 & 1.00 \\
 ResNet50 & DS-2 & 0.50 & 0.47 & 0.38 & 0.62 & 0.50\\
  & DS-3 & 0.50 & 0.32 & 0.24 & 0.49 & 0.59\\ 
  \midrule
  & DS-1 & 1.00 & 1.00 & 1.00 & 1.00 & 1.00  \\
 MobileNet & DS-2 & 0.50 & 0.47 & 0.38 & 0.62 & 0.50\\ 
  & DS-3 & 0.50 & 0.32 & 0.24 & 0.49 & 0.50\\ 
\bottomrule             
\end{tabular}
}
\end{table}





\begin{figure}[!t]
\captionsetup{justification=centering}
\centerline{\includegraphics[width = \columnwidth]{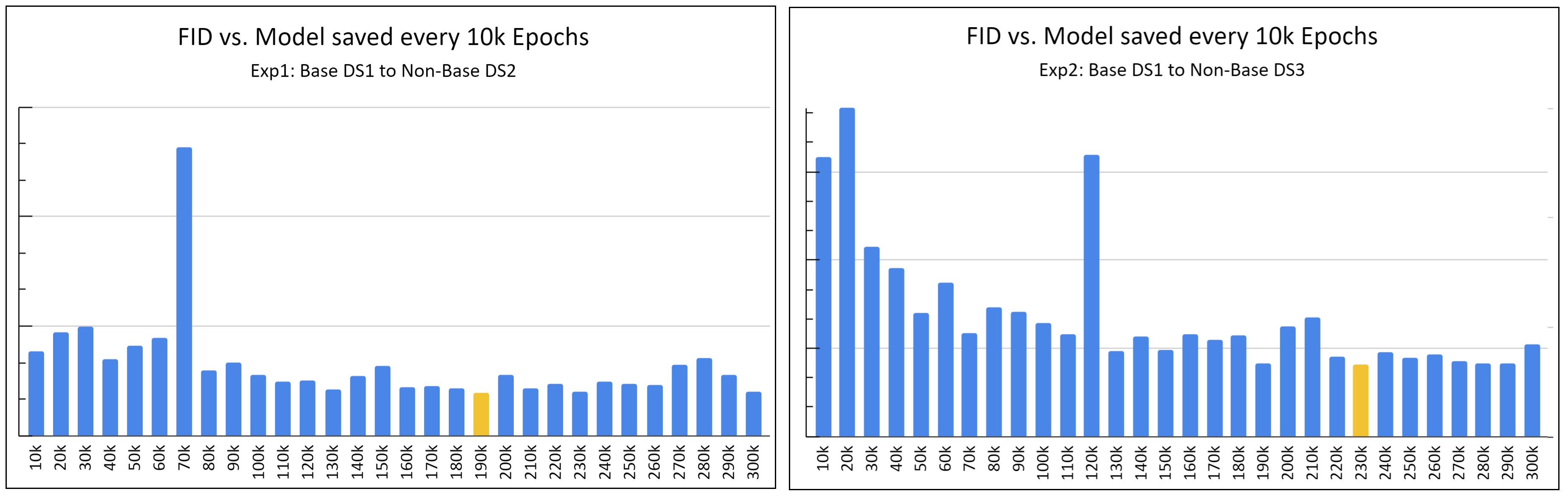}}
\caption{\capitalisewords{FID Values for Images Generated from the Validation Sets of DS-2 and DS-3 Using the 30 saved UI2I translation checkpoint models.}}
\label{fig5}
\end{figure}

{
\subsection{Experiment II: Source Classification Model on Target Images Translated Using UI2I Translation}
In this experiment, we first apply the UI2I translation model to translate images from the target DS which is then used as input for the source classification model for classification. Once the UI2I translation model started training, a checkpoint model was saved every $10k$ iterations until the end of the training session which was set to $300k$ iterations. Thus, a total of $30$ models were saved. Due to the absence of labels in unsupervised machine learning, conventional supervised evaluation metrics cannot be employed to determine the optimal stopping point for model training. Thus, each of these 30 models was evaluated on the target validation dataset using the unsupervised FID metric. The model that scored the best FID metric (lowest value), was selected to be used in rendering generated images from the target test dataset. Fig\ref{fig5} shows the FID values for images generated from the target validation sets of DS-2 and DS-3 using the 30 saved checkpoint models.
}

As seen in Fig\ref{fig5}, the models saved at epochs 190k and 230k were the best FID scoring models for DS-2 and DS-3 respectively. The best FID scoring model was then applied to the target testing data set to translate them into the source DS domain. The target test set images translated by the best FID scoring model were classified using the source classification model. The results were then compared with the evaluation results when using the source classification model on the same target test set without translation to showcase the efficiency of the proposed framework. The results of this comparison are summarized in Table\ref{table3}. As seen from the table, the models performance are improved on all five evaluation metrics. 

\begin{table}
\captionsetup{justification=centering}
\caption{\capitalisewords{Source Classification Models Evaluations on DS-2 and DS-3 With and Without Translation.} }
\label{table3}
\centering
\resizebox{\columnwidth}{!}{%
\begin{tabular}{cccccc}
\toprule
 & & \multicolumn{2}{c}{\textbf{DS-2}} & \multicolumn{2}{c}{\textbf{DS-3}} \\
\cmidrule(lr){3-6}
\textbf{Classifier} & \textbf{Metric} & \textbf{w/o translation} & \textbf{with translation} & \textbf{w/o translation} & \textbf{with translation} \\
\midrule
 & Balanced Accuracy & 0.50 & 0.75 & 0.50 & 0.90 \\
 & F1 weighted & 0.47 & 0.71 & 0.32 & 0.90 \\ 
\textbf{Customized Model} & Precision weighted & 0.38 & 0.80 & 0.24 & 0.90\\
 & Recall weighted & 0.62 & 0.71 & 0.49 & 0.90\\ 
 & ROC AUC & 0.37 & 0.77 & 0.89 & 0.95\\ 

\midrule

 & Balanced Accuracy & 0.50 & 0.66 & 0.50 & 0.92 \\
 
 & F1 weighted & 0.47 & 0.69 & 0.32 & 0.92 \\ 
 
\textbf{ResNet} & Precision weighted & 0.38 & 0.69 & 0.24 & 0.92\\

 & Recall weighted & 0.62 & 0.69 & 0.49 & 0.90\\ 
 
 & ROC AUC & 0.50 & 0.73 & 0.59 & 0.97\\
 
 \midrule
 
 & Balanced Accuracy & 0.50 & 0.63 & 0.50 & 0.89 \\ 
 
 & F1 weighted & 0.47 & 0.66 & 0.32 & 0.89 \\ 
 
\textbf{MobileNet} & Precision weighted & 0.38 & 0.66 & 0.24 & 0.90 \\

 & Recall weighted & 0.62 & 0.66 & 0.49 & 0.89 \\ 
 
 & ROC AUC & 0.50 & 0.68 & 0.50 & 0.96 \\ 

\bottomrule             
\end{tabular}
}
\end{table}

\begin{figure}[!t]
\captionsetup{justification=centering}
\centerline{\includegraphics[width=\columnwidth]{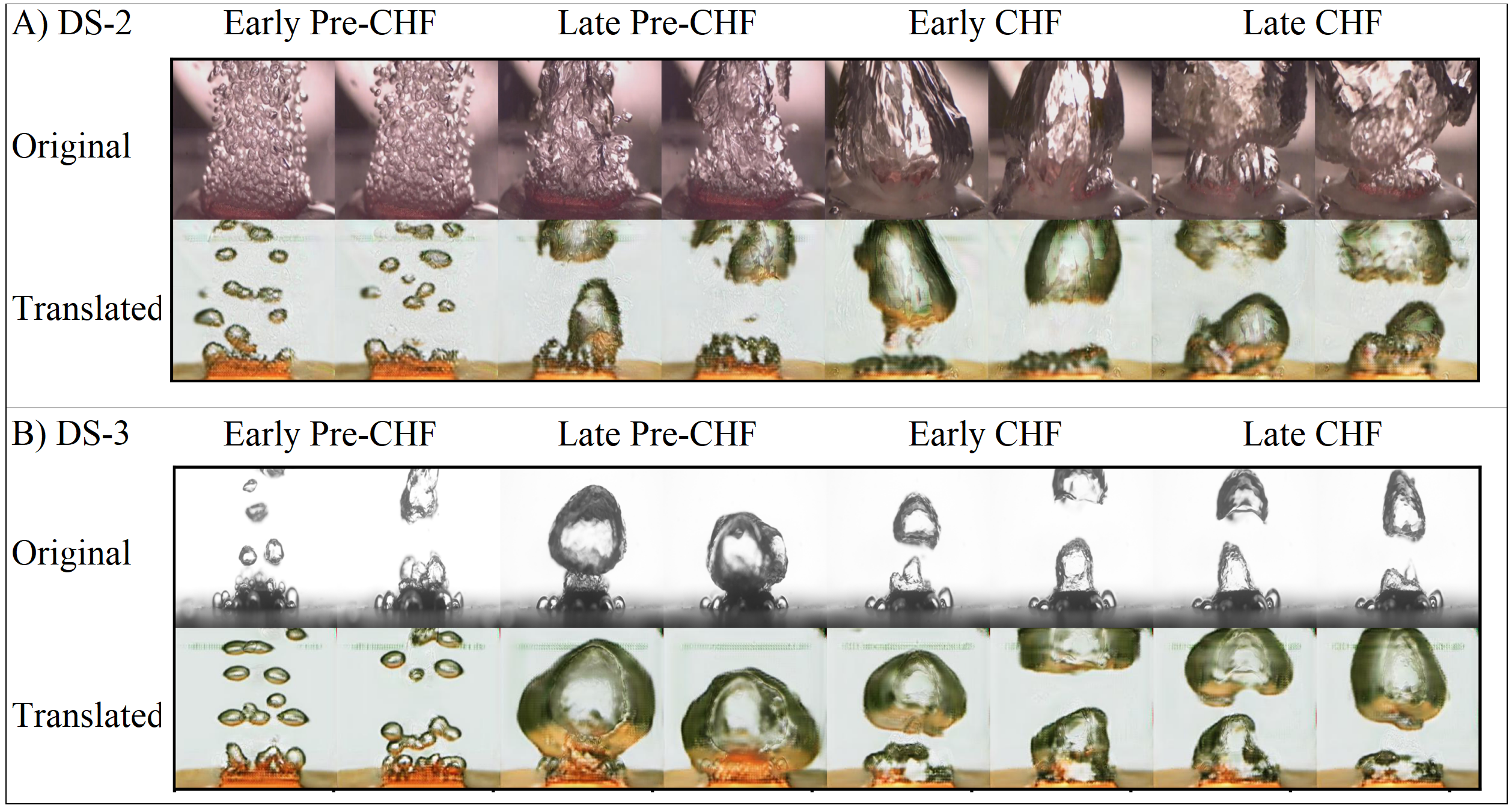}}
\caption{\capitalisewords{{Samples Generated by the best FID Scoring Models for A) DS-2 $\xrightarrow{}$ DS-1 and B) DS-3 $\xrightarrow{}$ DS-1.}}}
\label{fig6}
\end{figure}

{
\subsection{ Discussion on Framework Efficacy in Boiling Crisis Detection}}
Fig \ref{fig6} shows samples generated from each class for DS-2 and DS-3, respectively. The first row shows samples from the real target DS images, while the second row shows the translated version of the same image generated using the best FID scoring model.

The confusion matrices {generated using the custom-built source classification model} for both datasets are presented in Table\ref{table4}. As seen, for DS-2, the model was able to predict 118 out of 124 CHF (95.2\%) correctly and 111 out of 200 (55.5\%) Pre-CHF. For DS-3, the model was able to predict 1109 out of 1216 CHF (91.2\%) correctly and 1040 out of 1172 (88.7\%) Pre-CHF. We believe this initial effort shows the promise of using our framework in generalizing boiling crisis detection models.

\begin{table}
\captionsetup{justification=centering}
\caption{\capitalisewords{Confusion Matrices: Custom-Built Source Classification Model Predictions on Translated Images from DS-2 and DS-3.}}
\label{table4}
\centering
\resizebox{\columnwidth}{!}{%
\begin{tabular}{lccccc}
\toprule
 & & \multicolumn{2}{c}{\textbf{Source DS-1 to  Target DS-2}} & \multicolumn{2}{c}{\textbf{Source DS-1 to  Target DS-3}} \\
\cmidrule(lr){3-6}
 & & \multicolumn{2}{c}{\textbf{Predicted}} & \multicolumn{2}{c}{\textbf{Predicted}} \\
\cmidrule(lr){3-6}
 & & \textbf{CHF} & \textbf{Pre-CHF} & \textbf{CHF} & \textbf{Pre-CHF} \\
\midrule
\multirow{2}{*}{\textbf{True}}& \textbf{CHF} & 118 & 6 & 1109 & 107  \\
 & \textbf{Pre-CHF} & 89 & 111 & 132 & 1040 \\

\bottomrule             
\end{tabular}
}
\end{table}

\begin{figure}[!t]
\captionsetup{justification=centering}
\centerline{\includegraphics[width=\columnwidth]{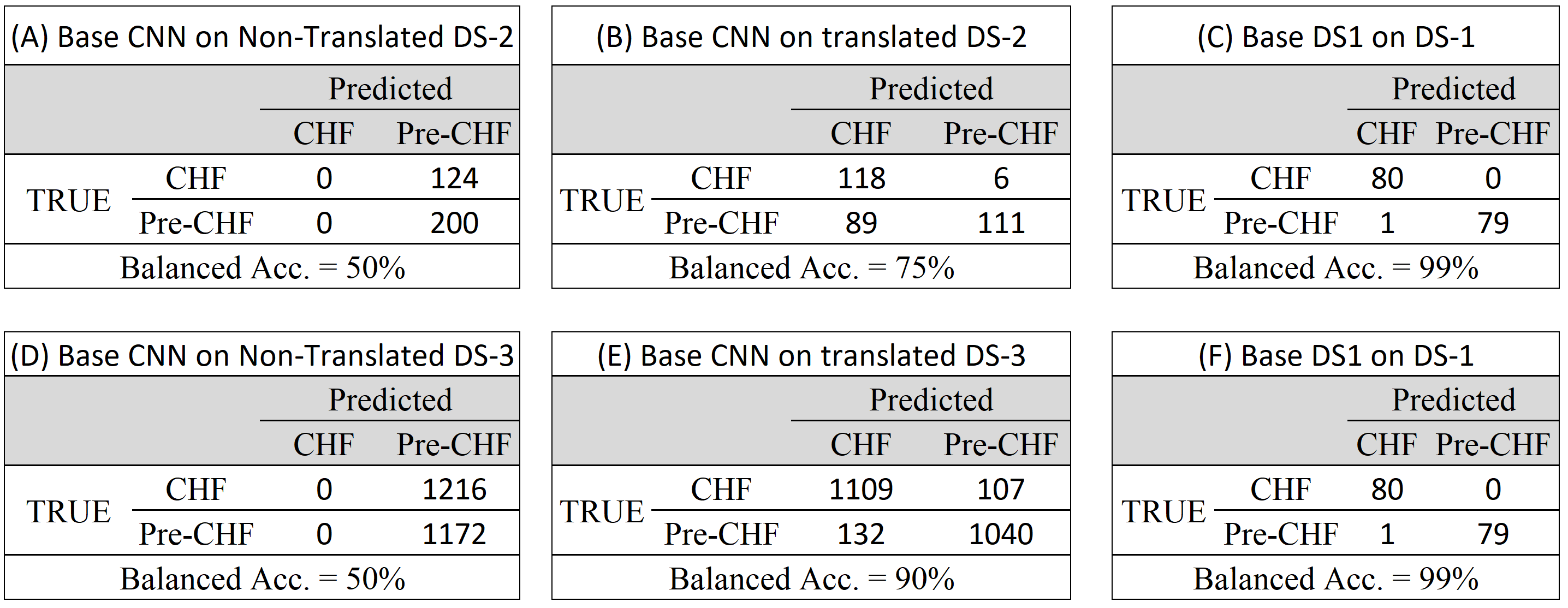}}
\caption{\capitalisewords{Comparison of All Confusion Matrices Generated by The Custom-Built Source Classification Model.}}
\label{fig7}
\end{figure}

For a more in-depth comparative analysis of the results, {a comparison of all confusion matrices generated by the custom-built source classification model} are displayed in Fig\ref{fig7}. The first column (Tables A and D) shows the results of the model when tested blindly on target DS images without any translation (worst-case scenario). As expected, the results show that the model is not generalizable to foreign datasets and will produce results equivalent to randomly guessing with a balanced accuracy of 50\% for both DS-2 and DS-3. The second column (Tables B and E) shows the results when using our framework and how it significantly improves the generalization of the Base CNN model for both DS-2 and DS-3 than when used blindly as in the first column. It also shows how far away is our method from the best-case scenario shown in the third column (Tables D and F) where the Base CNN model was tested on the same dataset that it was previously trained on.

\begin{figure}[!t]
\captionsetup{justification=centering}
\centerline{\includegraphics[width = \columnwidth]{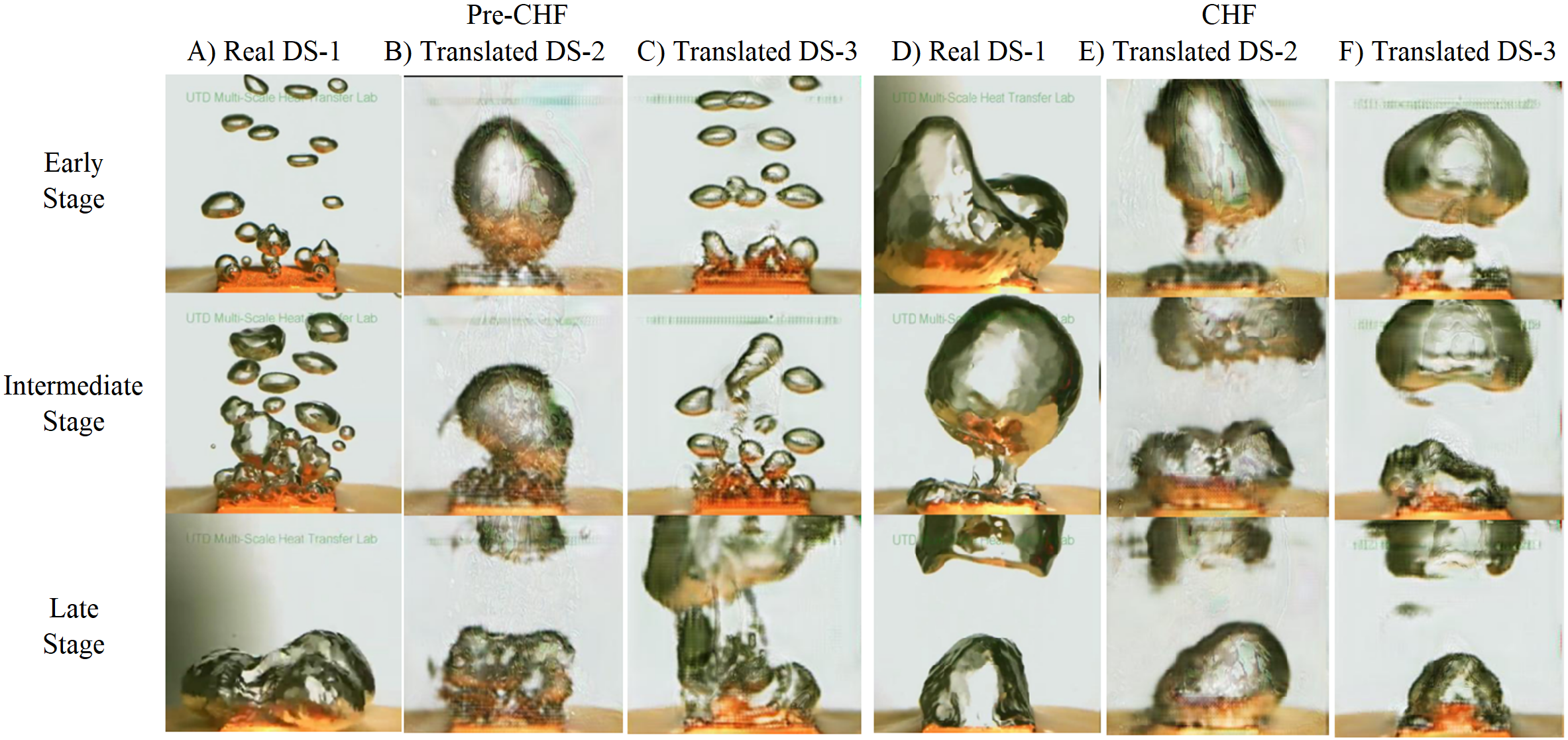}}
\caption{\capitalisewords{Samples from Real DS-1. Translated DS-2 and Translated DS-3 from Different Stages for Each Regime.}}
\label{fig8}
\end{figure}

It is observed that there is a significant improvement in the results when the images are first translated from the target DS to look like DS-1 images and then tested by the source classification model rather than blindly testing the model on the target DS images without any translation. All metrics have improved significantly, but most importantly, the “balanced accuracy” metric has increased by 50\% when using DS-2 as target, and by 80\% when using DS-3 as the target. This clearly demonstrates the effectiveness of the framework in translating the images to a domain that the classification model has seen before and how this translation will improve the results obtained by the same model on the same images before the translation.
To further understand the performance of the model, a comparison between samples from the real and the generated images for each class during different time intervals was plotted for each experiment. Fig\ref{fig8} A), B) and C) shows the comparisons for the “pre-CHF”, while D), E) and F) shows the comparisons for the “CHF” class. As seen when comparing Fig\ref{fig8} A) and B), the model for DS-2 doesn’t seem to be able to identify that the early and intermediate stages “pre-CHF” from DS-2 should be translated to early and intermediate -stages DS-1 “pre-CHF” images, but it has more success in translating the latter stages of DS-2 “pre-CHF” to look like their counterparts in DS-1 “pre-CHF”. This could be one of the reasons why the misclassifications percentage is higher for this class. Another reason could be attributed to the distortions (or fidelity) of the images.  The performance for DS-3 is much better for the “Pre-CHF” class. As seen when comparing Fig\ref{fig8} A) and C), the translated images have much better quality throughout the entire process and with little to no traces of the original domain appearing in most images. The model also seems successful in allocating most images to their proper timeline as in the real images which explains the superiority of the results of DS-3 over DS-2 for this class.
The performance of the model in the “CHF” class for DS-2 was much better than the “pre-CHF” class. The model seems to be able to translate each stage of this class to it’s correct counterpart. The translated images such as those viewed in Fig\ref{fig8} E) seem to be exhibiting the same distortion problem as in the case of the “pre-CHF” class. The performance on DS-3 for the “CHF” class is also similar to the DS-2 performance, as seen in Fig\ref{fig8} F); however, the translated DS-3 images seem to have much better quality throughout the entire process and with little to no traces of the original domain appearing in most images. The model also seems successful in allocating most images to their proper timeline as in the real images.

\begin{figure}[!t]
\captionsetup{justification=centering}
\centerline{\includegraphics[width=\columnwidth]{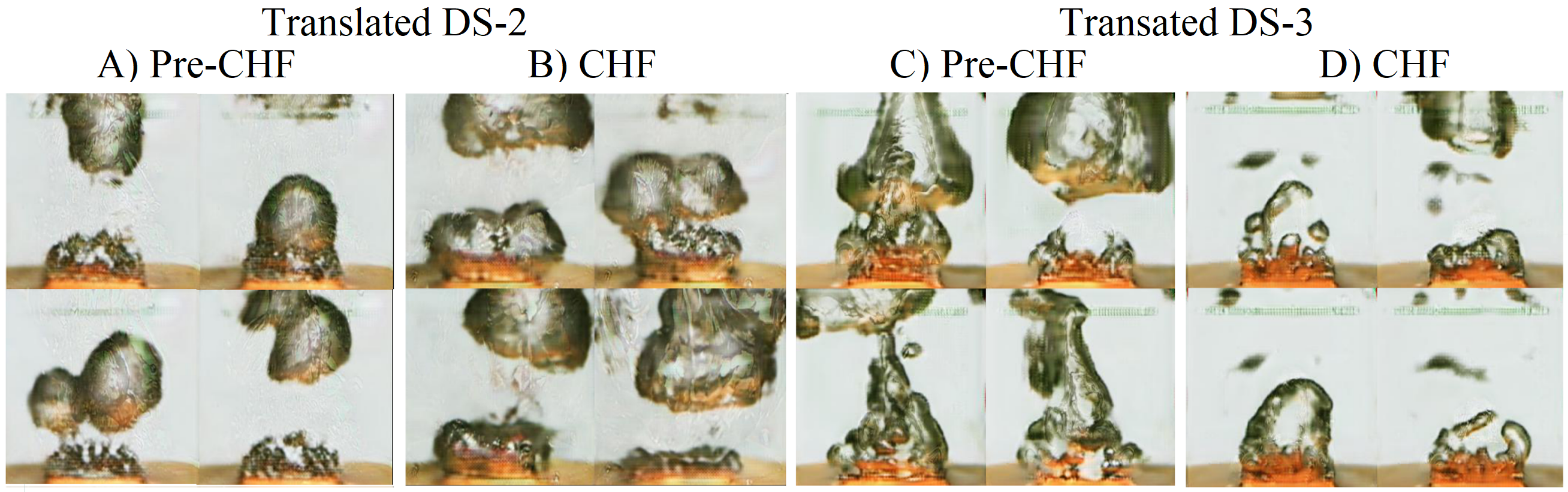}}
\caption{\capitalisewords{Examples of Misclassified Images from Each DS and Class.}}
\label{fig9}
\end{figure}

Fig\ref{fig9} A) and B) show a sample of the misclassified images for DS-2 for the “pre-CHF” and the “CHF” classes respectively. As observed, the major reason behind the misclassifications of the “pre-CHF” class for DS-2 is that the translated images look more like the “CHF” class rather than the “pre-CHF” class. The reason for the misclassifications in the“CHF” images for DS-2 could be that the six misclassified images suffer from distortion and immature translation effects. Similarly, Fig\ref{fig9} C) and D) show a sample of the misclassified images for DS-3. All classifications for the “pre-CHF” class are occurring exclusively in the later half of this stage (from 60W to 120W) where the images start to look similar to the “CHF” state as apparent in the figure. Misclassified images from both classes seem to be suffering from apparent immature translation artifacts that we suspect are the major cause for the confusion in the classification.

\subsection{Discussion on FID}
\begin{figure}[!t]
\captionsetup{justification=centering}
\centerline{\includegraphics[width = \columnwidth]{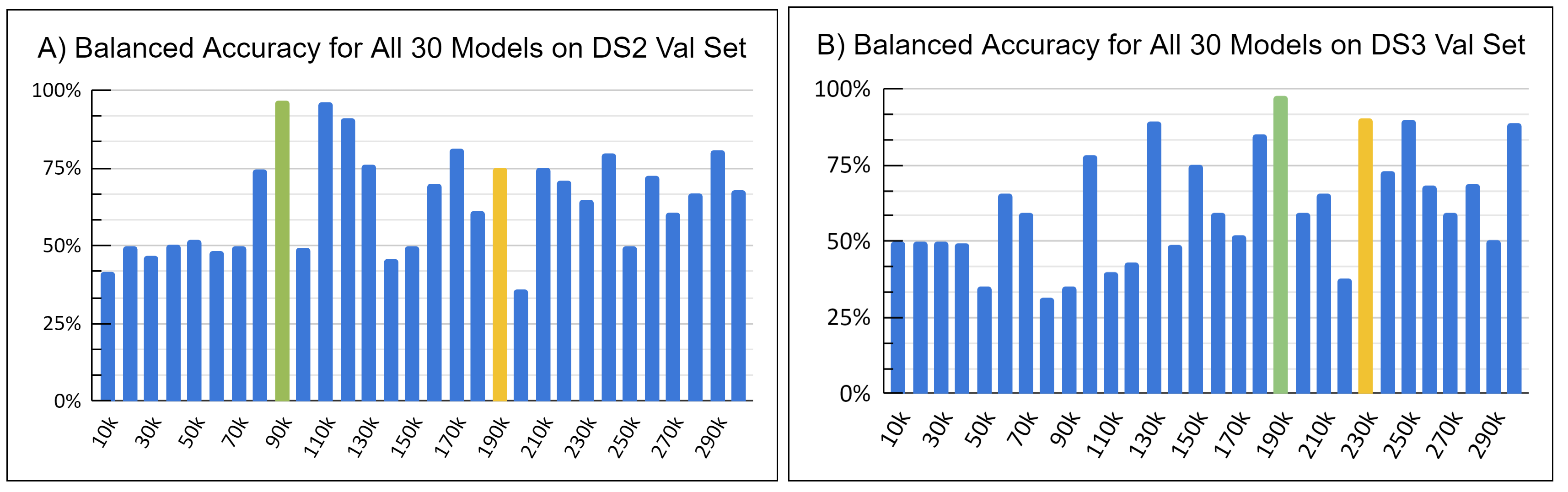}}
\caption{\capitalisewords{Metrics Comparison Between Best Attainable Models and FID Selected Models on Test Set for DS-2 and DS-3.}}
\label{fig10}
\end{figure}

How good is the FID metric in selecting the best saved UI2I translation model checkpoint? To test this, we assumed the labels for the validation data are available and used balanced accuracy instead of FID for model selection. Fig\ref{fig10} A) and B) show the balanced accuracy scores for every model on the labeled validation set for DS-2 and DS-3 respectively. For DS-2, out of the 30 models, the images generated by model 90k achieved the highest balanced accuracy when tested with the custom-built source classification model with a value of 96\% on the validation set compared to 75\% achieved by model 190k which was selected by the FID metric. The 90k model was then used to generate fake images from the test set and these images were also tested using the custom-built source classification model, achieving a balanced accuracy of 95\% as compared to the 75\% achieved when using the 190k model as seen in Table 5. For DS-3, out of the 30 models, the images generated by model 190k achieved the highest balanced accuracy when tested with the Base CNN model with a value of 98\% on the validation set compared to 90\% achieved by model 230k that was selected by the FID metric. The 190k model was then used to generate fake images from the test set and these images were also tested using the custom-built source classification model, achieving a balanced accuracy of 98\% as compared to the 90\% achieved when using the 230k model as seen in Table\ref{table5}.

\begin{table}
\captionsetup{justification=centering}
\caption{\capitalisewords{Confusion Matrices of Custom-Built Source Classification Model Predictions on Translated Images from DS-2 and DS-3.}}
\label{table5}
\centering
\resizebox{\columnwidth}{!}{%
\begin{tabular}{
l
>{\centering\arraybackslash}m{2.5cm}
>{\centering\arraybackslash}m{2cm}
>{\centering\arraybackslash}m{2cm}
>{\centering\arraybackslash}m{2cm}
>{\centering\arraybackslash}m{2cm}
>{\centering\arraybackslash}m{2cm}
}

\toprule
\textbf{DS} & \textbf{Model Name} & \textbf{Balanced Accuracy} & \textbf{F1 weighted} & \textbf{Precision weighted} & \textbf{Recall weighted} & \textbf{ROC AUC} \\
\midrule
\multirow{2}{*}{{DS-2}}& {90k model} & 0.95 & 0.96 & 0.96 & 0.96 & 0.99  \\
 & {190k model} &  0.75 & 0.71 & 0.80 & 0.71 & 0.77 \\
 \midrule
\multirow{2}{*}{{DS-3}}& {190k model} & 0.98 & 0.98 & 0.98 & 0.98 & 1.00 \\
 & {230k model} & 0.90 & 0.90 & 0.90 & 0.90 & 0.95 \\ 
\bottomrule             
\end{tabular}
}
\end{table}

\begin{figure}[!t]
\captionsetup{justification=centering}
\centerline{\includegraphics[width=\columnwidth]{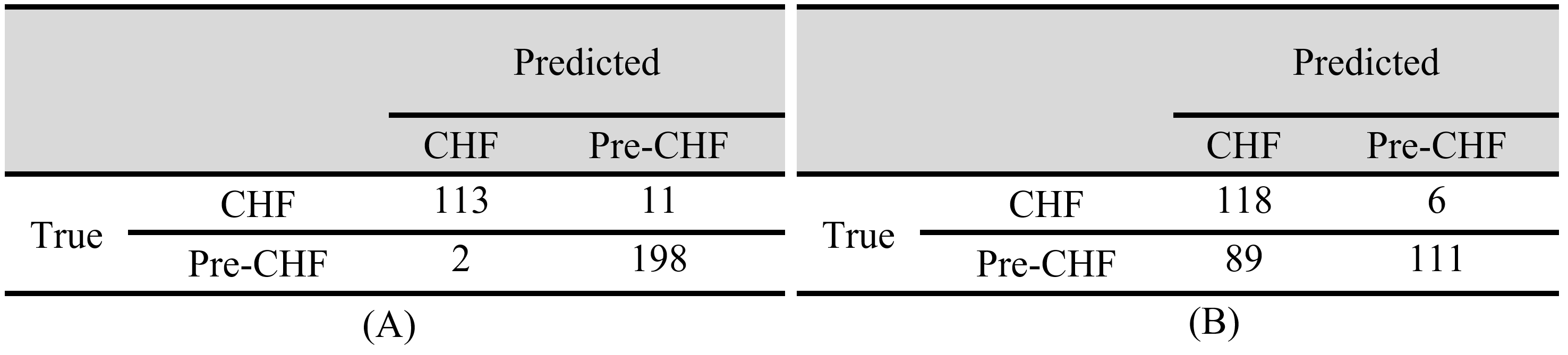}}
\caption{ \capitalisewords{Confusion Matrices for (A)90k and (B)190k Models Using The Customized Source Classification Model.}}
\label{fig11}
\end{figure}

The confusion matrices for both models for DS-2 are displayed in Fig\ref{fig11}. There is a notable improvement in the prediction accuracy of the “Pre-CHF” class, it has improved from 56\% to 99\%, while the predicting accuracy for the “CHF” class slightly decreased from 95\% to 91\%. 

\begin{figure}[!t]
\captionsetup{justification=centering}
\centerline{\includegraphics[width=\columnwidth]{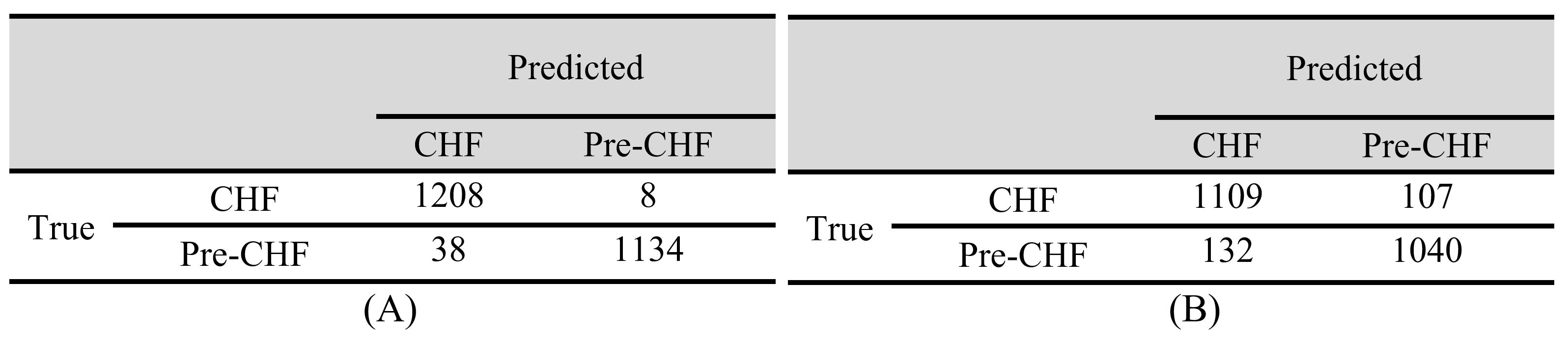}}
\caption{ \capitalisewords{Confusion Matrices for (A) 190k and (B) 230k Models Using The Customized Source Classification Model.}}
\label{fig12}
\end{figure}

The confusion matrices for both models for DS-3 are displayed in Fig\ref{fig12}. There is a notable improvement in the prediction accuracy of both classes. The accuracy for the “pre-CHF” class improved from 89\% to 97\%, while the predicting accuracy for the “CHF” class increased from 91\% to 99\%.

As seen from the results above, in general, although the FID metric provides a good solution, it may not guarantee the best attainable one. { In the future, we plan on developing a new unsupervised metric to support cross-domain classification applications such as this one. An example of a possible direction could be in utilizing distance metrics based on the principal components derived from the images \citep{rokoni2022a} as our next step.}  

{
\section{Conclusion}
This paper presents a framework that utilizes UI2I translation for generalized boiling crisis detection models. The approach involves training a classification model on boiling images from a source dataset for detecting boiling crisis. To classify boiling images from new datasets (target datasets), where they are first translated into the source domain using the UI2I translation model and then classified by the source classification model. Three distinct boiling datasets from different research groups were used to demonstrate the efficacy of the proposed framework.

The results show that The proposed framework enables the source classification model to yield higher performance in comparison to its performance without the framework. Moreover, the results show that the FID can help select a UI2I translation model checkpoint with reasonably good performance, but is inconsistent and unable to select the best attainable model. Future efforts are needed to be directed toward developing more reliable metrics for selecting the best attainable UI2I translation model checkpoint.

Compared to existing visualization-based CHF detection studies that are predominately based on datasets with labels, the presented work demonstrates the feasibility of classifying boiling images from any new datasets without labels. This work presents a step forward towards generalizing classification models for engineering applications and making visualization-based boiling crisis a viable monitoring and detection tool in industrial applications. Moreover, The proposed framework could be generalized to work with any classification model (not just CNN) using any UI2I translation model (not just FP-GAN). Finally, the proposed framework has the capability to be utilized in any other similar applications where a pre-trained classifier needs to be generalized to accommodate unlabeled datasets from foreign domains. A future direction for this work is to explore further this capability and demonstrate its efficacy in applications other than CHF detection. 
}

\section*{CRediT authorship contribution statement}
\textbf{Firas Al-Hindawi:} Conceptualization, Methodology, Software, Writing - Original Draft, Writing - Review \& Editing. \textbf{Tejaswi Soori:} Data Curation, Writing - Original Draft, Writing - Review \& Editing. \textbf{Han Hu:} Conceptualization, Writing - Original Draft, Writing - Review \& Editing, Data Curation. \textbf{ Md Mahfuzur Rahman Siddiquee:} Conceptualization, {Methodology,} Software, Writing - Original Draft, Writing - Review \& Editing. \textbf{Hyunsoo Yoon:} Conceptualization, Writing - Review \& Editing, Supervision. \textbf{Teresa Wu:} Conceptualization, Methodology, Writing - Original Draft, Writing - Review \& Editing, Resources, Supervision, Project administration. \textbf{Ying Sun:} Writing - Review \& Editing, Supervision.

\section*{Declaration of Competing Interest}
The authors declare that they have no known competing financial interests or personal relationships that could have appeared to influence the work reported in this paper

\section*{Acknowledgement}
Support for this work was provided in part by the US National Science Foundation under Grant No. CBET-1705745.

\bibliography{citations.bib}

\end{document}